\documentclass{article}

    \usepackage[final]{neurips_2020}

\usepackage{natbib}
\setcitestyle{square,sort,comma,numbers}
\usepackage[utf8]{inputenc} 
\usepackage[T1]{fontenc}    
\usepackage{hyperref}       
\usepackage{url}            
\usepackage{booktabs}       
\usepackage{amsfonts}       
\usepackage{nicefrac}       
\usepackage{microtype}      
\usepackage{graphicx}
\graphicspath{{images/}}
\usepackage{amsmath}
\newcommand{\R}{\mathbb{R}}
\usepackage{caption} 
\usepackage[toc,page,header]{appendix}

\usepackage{tablefootnote}
\usepackage{multicol}
\usepackage{caption}
\usepackage{amsmath}
\title{Self-Supervised Learning for Fine-Grained Visual  Categorization}

\author{
    Muhammad Maaz \\
   \texttt{20020063@mbzuai.ac.ae} \\
   \And
   Hanoona Abdul Rasheed \\
   \texttt{20020070@mbzuai.ac.ae} \\
   \And
   Dhanalaxmi Gaddam \\
   \texttt{20020011@mbzuai.ac.ae} \\
}

\begin{document}

\maketitle
\begin{abstract}
Recent research in self-supervised learning (SSL) has shown its capability in learning useful semantic representations from images for classification tasks. Through our work, we study the usefulness of SSL for Fine-Grained Visual Categorization (FGVC). FGVC aims to distinguish objects of visually similar sub categories within a general category. The small inter-class, but large intra-class variations within the dataset makes it a challenging task. The limited availability of annotated labels for such a fine-grained data encourages the need for SSL, where additional supervision can boost learning without the cost of extra annotations. Our baseline achieves $86.36\%$ top-1 classification accuracy on CUB-200-2011 dataset by utilizing random crop augmentation during training and center crop augmentation during testing. In this work, we explore the usefulness of various pretext tasks, specifically, rotation, pretext invariant representation learning (PIRL), and deconstruction and construction learning (DCL) for FGVC. Rotation as an auxiliary task promotes the model to learn global features, and diverts it from focusing on the subtle details. PIRL that uses jigsaw patches attempts to focus on discriminative local regions, but struggles to accurately localize them. DCL helps in learning local discriminating features and outperforms the baseline by achieving $87.41\%$ top-1 accuracy. The deconstruction learning forces the model to focus on local object parts, while reconstruction learning helps in learning the correlation between the parts. We perform extensive experiments to reason our findings. Our code is available on \href{https://github.com/mmaaz60/ssl_for_fgvc}{GitHub}\footnote{https://github.com/mmaaz60/ssl\_for\_fgvc}.
\end{abstract}
\section{Background}


Today, deep learning models have achieved very inspiring results in several fields  \cite{he2016deep, deng2009imagenet}. Although we humans can perfectly learn to recognize a set of categories with very few samples of data, convolutional neural networks need to be trained on a huge number of examples per class in order to distinguish between a set of classes. Therefore, the performance of these models greatly depends on the amount of annotated data available for training. However, collecting large scale annotated data is time consuming and expensive. 


Many SSL methods have been proposed as a solution to mitigate such challenging need for large scale annotated data. In this approach, the model instead learns visual features by learning objective functions of various pretext tasks applied on a limited available labelled data, in other words the model learns from large scale unlabelled data. SSL tasks mainly consists of two learning stages, learning the image representations with the help of some pretext self-supervised tasks and then adapting these representations for a downstream task; the actual task the model is designed for \cite{gidaris2019boosting}.

Fine-grained recognition focuses on distinguishing objects from different subordinate level categories within a general category \cite{wei2019deep}. The fine-grained recognition tasks are extremely challenging as the different children of a category would have only few subtle differences between them. The major goal in this problem setting is to learn these subtle features such that the model can differentiate between objects that are visually very similar, such as the breed of a dog or recognizing the different species of the same bird. Another challenging goal is that the model has to be invariant to the intra-class variations such as viewpoint, appearance or location of the object in the image. Distinguishing between objects in this scenario often implies focusing on details from coarser to finer levels, such as a beak of a bird that are discriminating features for the recognition task.

\begin{figure}[h]
\centering
\includegraphics[width=\linewidth]{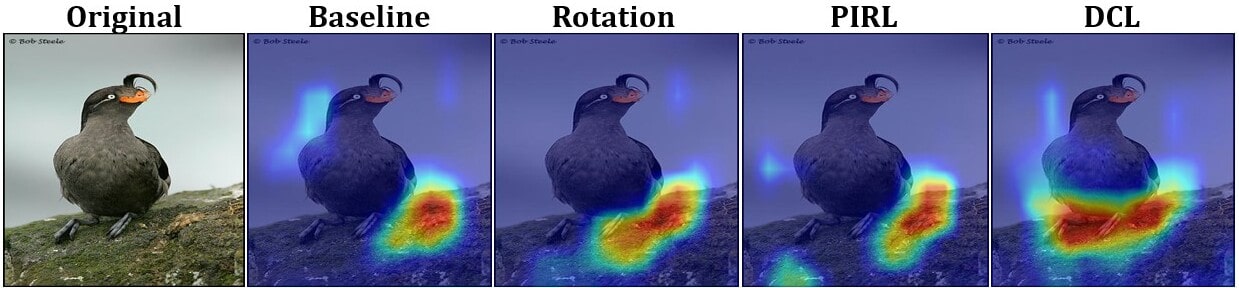}
\caption{The figure compares the Grad-CAM \cite{selvaraju2017grad} visualizations of the various auxiliary SSL tasks with the baseline on an image of a crested auklet. The DCL successfully localizes the discriminative features such as the bird's webbed feet and black claws.}
\label{fig:combined_cam}
\end{figure}

To study the effectiveness of SSL on FGVC, we draw inspiration for our architectural pipeline from \cite{gidaris2019boosting} and experimented with the various self-supervised auxiliary tasks on this. Specifically, the self-supervised loss is combined with the supervised loss in a parallel network to promote learning in the FGVC setting (Figure. \ref{rotation}). The various auxiliary tasks adopted for this work are rotation \cite{gidaris2018unsupervised}, PIRL \cite{misra2020self} and DCL \cite{chen2019destruction}.






\section{Related Work}
\subsection{Self-supervision} 
Self supervised pretraining is a recently emerging technique that aims to learn good representations from unlabelled image data, to relieve the dependency on large scale annotated data. It have proved to increase the efficiency and generalization capability of traditional models by enabling them to learn better features from visual data. Various pretext tasks have been applied for learning visual representations for improving classification tasks. Some common pretext tasks are based on spatial context structures such as predicting random rotations applied to image \cite{gidaris2018unsupervised}, jigsaw puzzle \cite{noroozi2016unsupervised}, or predicting the relative position of image patches \cite{doersch2015unsupervised}. With the training from these pretext tasks, the model learns to extract the important semantic features from the data that are then used for other downstream tasks. 

Other techniques such as contrastive learning involves learning transformation invariant feature representations of the data \cite{misra2020self, he2020momentum, zbontar2021barlow}. It aims at combining the similar samples of an image together, while the dissimilar samples are pushed away from each other. The selection of similar and dissimilar pairs is carried out by several approaches including, creating multiple augmented views \cite{wu2018unsupervised, ye2019unsupervised}, patches of same image \cite{hjelm2018learning, isola2015learning} and using different time-steps in video \cite{zhuang2020unsupervised,oord2018representation}. 

In our approach, we consider a multi-task setting, where we train a ResNet-50 backbone using joint supervision.

\subsection{Fine Grained Classification}
Recently, fine-grained classification has gained much attraction of the research community. Most existing works follow a pipeline of first localizing regions of the object and then describing discriminative object parts \cite{deng2013fine, yang2012unsupervised}. Some of the earliest approaches used object annotation for localizing the object and its parts \cite{chai2013symbiotic, yang2012unsupervised}. Similar methods in \cite{berg2013poof, xie2013hierarchical} used part annotations to extract more informative features. A major drawback of these approaches is that it is more challenging to collect part annotations than image labels. To avoid such costly annotation, several recent works focused on unsupervised or weakly-supervised part learning. A bank of convolutional filters was proposed in \cite{wang2018learning} to learn high quality discriminative patches with no extra annotation. \cite{lin2015bilinear} proposed an architecture that uses complex pooling methods. A spectral clustering on convolutional filters was proposed in  \cite{xiao2015application} to find representative filters for parts. \cite{tang2016learning} uses a weakly supervised strategy to find discriminative features and leverage them to perform the classification between similar instances using multi-instance learning. 

Other methods use self-supervision that learns to find inter-class discriminative features using a combination of networks as navigator, teacher and scrutinizer \cite{yang2018learning}. An optimal classifier on top of deep features was proposed in \cite{dubey2018pairwise}. To find more relevant features, \cite{FGR} adopt the diversification block (DB) which suppresses prominent discriminative regions in the class specific activation maps (CAMs) \cite{zhou2016learning}. It also uses gradient-boosting cross entropy (GCE) loss to focus on the confusing classes. More related to this work, \cite{singh2017hide} proposed a weakly-supervised hide and seek framework to localize only the most discriminative parts of an object rather than all relevant parts. The approach forces the network to look for other relevant parts by hiding the most discriminative parts in training.

\begin{figure}[h!]
\centering
\includegraphics[width=\linewidth]{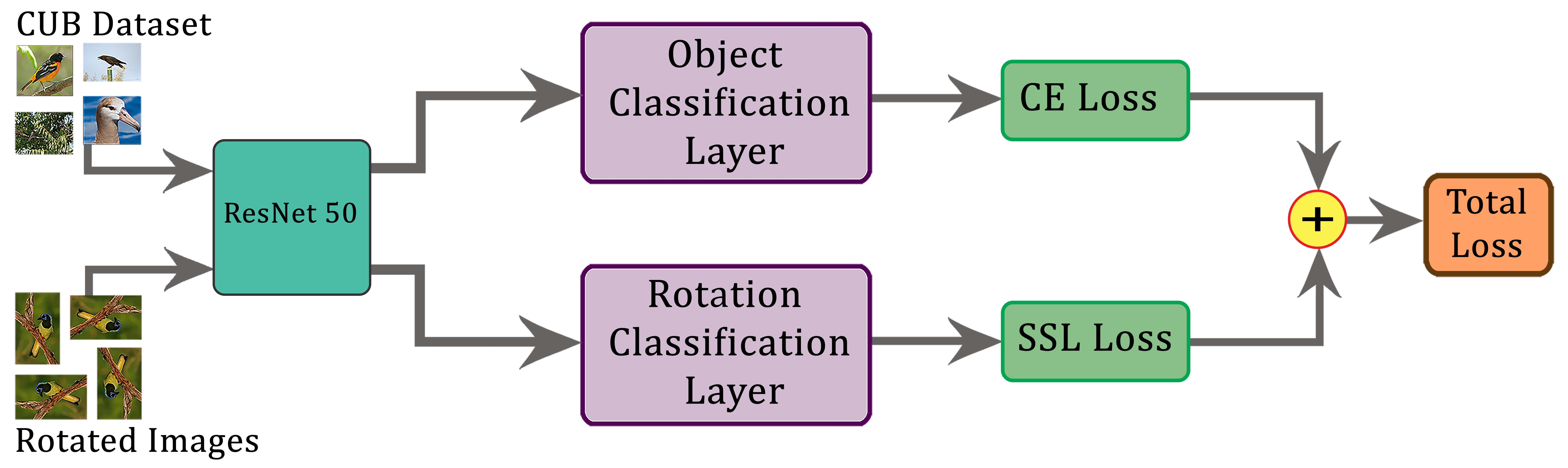}
\caption{The figure illustrates the architectural pipeline we follow throughout this work. We adopt a parallel pipeline design, where SSL loss is added to the supervised loss for end-to-end training. The features are extracted from the original and augmented images using a shared backbone. The network contains one head, each for supervised and self-supervised task. Here the auxiliary task is rotation.}
\label{rotation}
\end{figure}

\section {Methodology}
In our approach, we study the effectiveness of self-supervision as an auxiliary task for fine-grained image classification. We follow the concepts from \cite{gidaris2019boosting} and build a parallel network design for applying self-supervision to the CUB classification task. The original images are passed to ResNet-50 backbone followed by a classification head to calculate the class probability scores. The supervised cross entropy (CE) loss is calculated using these class probabilities. Alongside, the images are transformed to generate self-supervised labels for the SSL pretext task and are passed to the same ResNet-50 encoder. The SSL loss is calculated using the auto-generated labels and added to the supervised loss to compute the total loss. The complete pipeline is shown in  Figure. \ref{rotation}.

\subsection{Baseline Formation}
In order to experiment the effectiveness of various self-supervision tasks on the FGVC problem, we explored different ideas to build a strong baseline. Specifically, we experiment with a standard ResNet-50 and examine the architecture proposed in \cite{FGR}. The paper introduces DB and GCE loss to improve the FGVC performance by forcing the model to focus on the subtle discriminating features. To implement the proposed design, the global average pooling and the classification layer in the standard ResNet-50 are replaced with a $1 \times 1$ convolution layer with output channels equal to the number of classes. These class activation maps (CAMs) serve as the input to the diversification block. The details about DB and GCE are discussed in the appendix \ref{DB} and \ref{GCE}.



We found that the standard ResNet-50 model outperforms \cite{FGR} with DB and GCE in the same experimental settings. Therefore, our baseline consists of a standard ResNet-50 network. More details on the baseline formation are discussed in the experiments section.

\subsection{Rotation as SSL Pretext Task} 
We apply the concepts from \cite{gidaris2019boosting} on our baseline and study the effectiveness of the self-supervision as an auxiliary task for the fine-grained classification. The self-supervised rotation loss is introduced to extract more meaningful features from the images. The adopted pretext task is predicting the rotation transformations \cite{gidaris2018unsupervised} applied to the images, where each image is rotated by four multiples of 90 degrees $[0, 90, 180, 270]$. The model predicts the applied rotation along with the main fine-grained classification task. Equation. \ref{ssl_loss} shows the calculation of the total loss for the training, where $\lambda$ is a scale factor which decides the contribution of the rotation loss to the overall training loss.
\begin{equation} \label{ssl_loss}
    \mathcal{L}_{total} = (1 - \lambda) \mathcal{L}_{classification} + \lambda \ \mathcal{L}_{rotation}
\end{equation}

\subsection{Contrastive Self-supervised Learning} 
Contrastive Learning (CL) based SSL methods involve learning transformed invariant representations by maximizing the similarity between the transformed versions of the same image and maximizing the difference between the transformed versions of different images. These methods have been gaining popularity on account of their simplicity and exceptional results on many downstream tasks \cite{zbontar2021barlow, grill2020bootstrap, caron2020unsupervised}. In this work, we study the efficiency of these methods for the FGVC task on CUB dataset. We try different initialization for the backbone network of our baseline generated using contrastive self-supervised pre-training. Following that, we introduce contrastive self-supervised loss in a parallel fashion to the supervised network and study the usefulness of it on the FGVC task.

 \begin{figure}[h!]
 \centering
 \includegraphics[width=\linewidth]{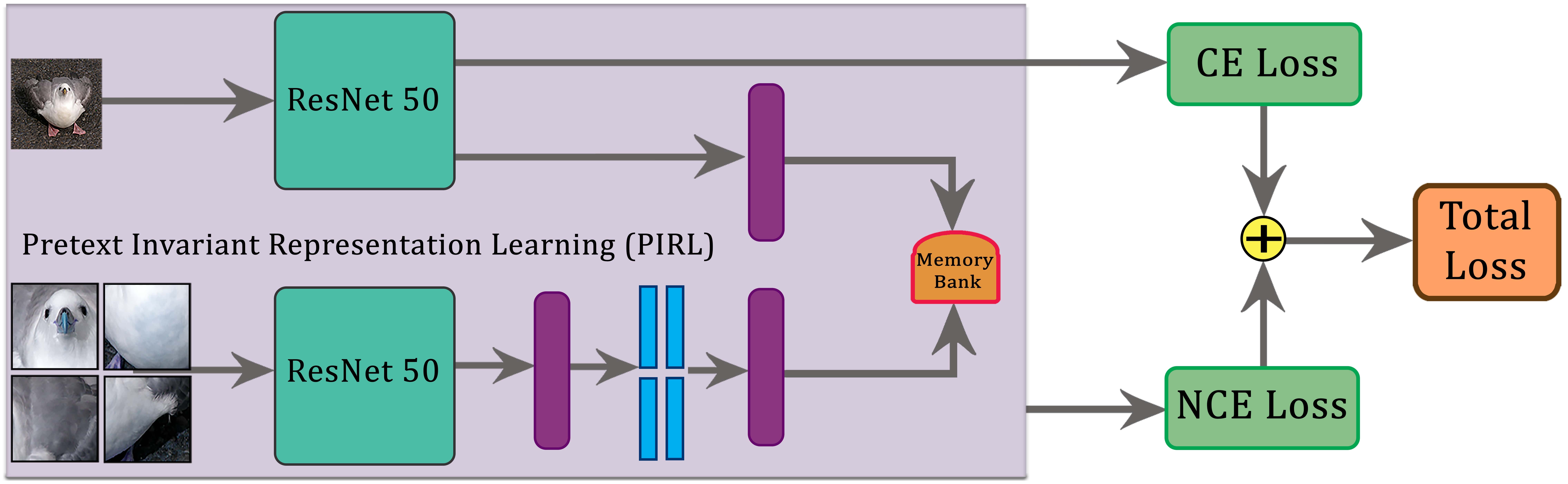}
 \caption{The figure illustrates the architectural pipeline of adding PIRL as an auxiliary task to the baseline. Original image together with the jigsaw patches from the same image are passed to shared ResNet-50 backbone to generate feature representations for calculating the contrastive self-supervised loss. The total loss is calculated by adding supervised and self-supervised loss.}
 \label{PIRL}
 \end{figure}
 
\paragraph{PIRL:}
Pretext-Invariant Representation Learning (PIRL) \cite{misra2020self} demonstrates that it is critical to learn transformation-invariant representations since visual semantics are not affected by image transformations. It proposes asymmetric networks to avoid learning trivial constant representation, which converges to a constant mapping for each input. The input image is passed through an encoder followed by a linear projection head that outputs a representation. Simultaneously, nine jigsaw \cite{noroozi2016unsupervised} patches are created from the original image and are passed to the same encoder to generate feature vectors. These features are concatenated and passed through another linear projector to obtain another representation. The noise contrastive estimator (NCE) computes the loss using these two representations as a positive pair, and considers representations in the memory bank as negative samples. The PIRL loss is given in Equation. \ref{PIRL_Loss}. 
\begin{equation} \label{PIRL_Loss}
    \mathcal{L}_{NCE}(I, I^t) = -\log[h(f(v_I), g(v_{I^t}))] - \sum_{I' \in D_N} \log[1 - h(g(v_{I^t}), f(v_{I'}))]
\end{equation}

Where the NCE term $h(v_I, v_{I^t})$ is defined as,

\begin{equation}
    h(v_I, v_{I^t}) = \frac{\exp\left( \frac{s(v_I, v_{I^t})}{\tau} \right )}
    {\exp\left( \frac{s(v_I, v_{I^t})}{\tau} \right ) + \sum_{I' \in D_N} \exp\left( \frac{s( v_{I^t}, v_{I'})}{\tau} \right )}
\end{equation}

Here, $D$ denotes the whole dataset and $D_N$ are the negative samples drawn from the memory bank excluding image $I$. $s()$ denotes the cosine-similarity between the two encoded representations and $\tau$ is a temperature hyper-parameter. Note that Equation. \ref{PIRL_Loss} maximizes the similarity between the representation of the original image ($I$) and the representation of the transformed image ($I^t$), whilst minimizing the similarity between $I$ and the representation of negative samples ($I'$).

PIRL uses memory bank to store the feature representations and hence does not require large batch size to provide large number of negative samples to the loss in Equation. \ref{PIRL_Loss}. The memory bank stores the representation $m_i$ of each image $I_i$ in the dataset $D$ where $m_i$ is the exponential moving average of the representations computed in the earlier epochs for the same image $I_i$. 


\subsection{Deconstruction and Construction Learning (DCL)}
To attain high performance in a fine-grained classification setting, the model must learn to identify and pay attention to local parts and learn discriminative feature representations. SSL that forces the model to look at smaller details from local regions like object parts, would therefore be a good fit for fine-grained classification. The paper \cite{chen2019destruction} proposes an interesting approach to learn such fine details by deconstruction and construction of the images by shuffling local regions, similar to jigsaw puzzles. This is done by first deconstructing the global structure of the images, in order to force the model to look at local object parts, and then training the model to learn to reconstruct the image to learn the semantic relations between the parts. The images are carefully deconstructed using a region confusion mechanism (RCM) and is used to train the model in a self-supervised fashion. The model is trained with three loss functions; a classification loss ($\mathcal{L}_{cls}$) to capture local object parts, an adversarial loss ($\mathcal{L}_{adv}$) that helps the model to distinguish the jigsaw images from the original, and lastly a reconstruction location loss ($\mathcal{L}_{loc}$) to learn the semantic correlation between the local regions.
\begin{equation}
\mathcal{L} = \mathcal{L}_{cls} + \mathcal{L}_{adv}+ \mathcal{L}_{loc}
\end{equation}

\begin{figure}[h!]
\centering
\includegraphics[width=\linewidth]{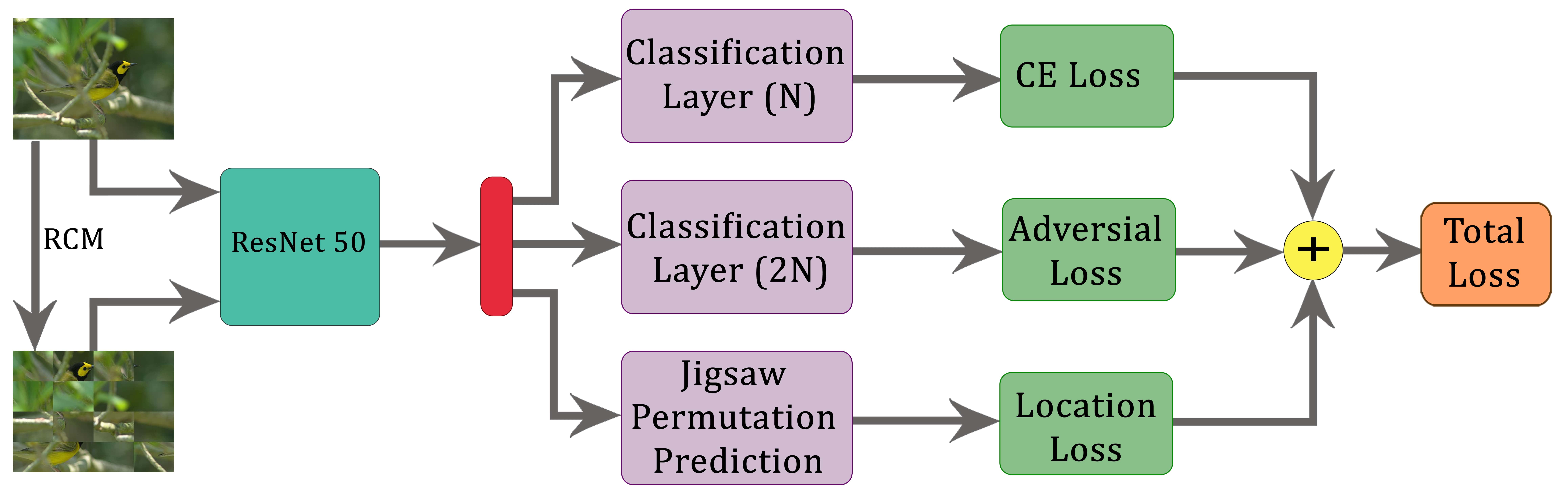}
\caption{The figure illustrates the architectural pipeline for the DCL. The image is carefully deconstructed using RCM and forwarded to a shared backbone together with the original image for feature extraction. The network consists of three heads corresponding to the three loss functions; a classification loss, an adversarial loss, and a reconstruction location loss. The loss functions are combined and the model is trained end-to-end.}
\label{DCL}
\end{figure}
\paragraph{Deconstruction:}
The proposed RCM \cite{chen2019destruction} deconstructs the global structure of the images into local patches of uniform size. The images are first uniformly sampled into patches of $k \times k$ dimensions denoted by $R_{i,j}$. For every patch in the image, a permutation $\sigma(i,j)$ is generated, which is controlled by a hyper-parameter of distance range $D$. Here, $D$ defines the maximum neighbourhood range to which the patch may be shuffled to, such that $[1< D<k]$. To get the permutations in the rows, a random vector $p_{j}$ is generated for every $j$ of $R_{i,j}$. The $i^{th}$ element of $p_{j}$ is computed as $p_{j,i}= i + d$, where $d \sim U(-D, D)$, a random variable from a uniform distribution in $[-D, D]$. The permutation for the rows, $\sigma_j^{row}$ of each patch is then generated by sorting the vector $p_{j}$. Similarly, the permutation for the columns, $\sigma_i^{col}$ are computed, such that it satisfies the condition $\forall \ i \in {[1, \cdots, k]}$,
$$ |\sigma_j^{row}(i)-i| < 2D| \ \  \textrm{and} \ \  |\sigma_i^{col}(i)-i| < 2D$$

\paragraph{Classification:}
The model consists of two classification heads (Figure. \ref{DCL}), one which predict the $N$ classes of the FGVC problem and another that predicts the $2N$ classes, $N$ for non deconstructed and $N$ for deconstructed images. Both classification heads receive the feature embeddings extracted using a ResNet-50 backbone. The first classification head learns to predict the actual class from both the original ($I$) and the deconstructed image ($\phi (I)$). It maps the input images into classification probability scores $C(I, \phi (I))$. A classification loss is applied at this head such that the model will learn to recognize both the original and deconstructed images. This loss intuitively functions as self-supervision, because the model is forced to learn local discriminative parts that distinguish a class from other classes, using self generated additional training samples. The loss is given in Equation. \ref{lcls}, where $l$ represents the target labels.
\begin{equation}
    \mathcal{L}_{cls} = -\sum_{I \in \mathcal{I}} \mathbf{\textit{l}} \times \log [C(I) C(\phi (I))]
\label{lcls}
\end{equation}

\paragraph{Adversarial Learning:}
The deconstruction of the images with RCM may introduce some visual noise patterns within the feature space and care must be taken that the model do not learn to overfit to these noises. Furthermore, the model must be capable of differentiating the jigsaw images ($\phi (I)$) from the original images ($I$), and this is done by applying an adversarial loss at the second classification head. This can be intuitively thought of as a discriminator network, which identifies whether the image is deconstructed or not. The discriminator can be represented as, 
\begin{equation}
    D(I,\theta_{adv}) = \textit{softmax} (\theta_{adv} C(I,\theta_{cls}^{[1,m]}))
\end{equation}
Here, $\theta_{adv}$ are parameters of the discriminator network. $\theta_{cls}^{[1,m]}$ are parameters of the classification network from layer 1 to $m$. $C(I, \theta_{cls}^{[1,m]})$ is the feature embedding output from the $m^{th}$ layer of the network. The adversarial loss ($\mathcal{L}_{adv}$) applied at the discriminator can be expressed as shown in Equation. \ref{adv}. For the destructive learning in DCL, both $\mathcal{L}_{cls}$ and $\mathcal{L}_{adv}$ will contribute to enhance the discriminative local features by introducing self-supervision.
\begin{equation}
      \mathcal{L}_{adv} = - \sum_{I \in \mathcal{I}} \mathbf{\textit{c}} \times \log [D(I)] +  \mathbf{\textit{(1-c)}} \times \log [D(\phi (I))]
\label{adv}
\end{equation}

\paragraph{Reconstruction Learning:}
The model learns to focus on the local discriminative regions with the combination of classification and adversarial loss. However, in order to train the model to learn about the correlation of the different local regions, an additional SSL is included through reconstruction of the jigsaw images. The feature embedding from the shared backbone are processed by a $1 \times 1$ convolution layer, which is then passed through a $\tanh$ function to generate a mapping of size $k \times k$, where each value corresponds to the prediction of the patch location. The predictions can be represented as $M_{(i,j)}(I)$ and $M_{\sigma \ (i,j)}(\phi (I))$ for the jigsaw images. The reconstruction loss ($\mathcal{L}_{loc}$) is a mean square error between the original patch location coordinates and the predicted locations from reconstruction, which is expressed as,
\begin{equation}
      \mathcal{L}_{loc} = \frac{1}{N}\ \left [ \sum_{i=1}^{N} {\left (M_{\sigma \ (i,j)}(\phi (I)) - \begin{bmatrix}
                            i \\
                            j
                        \end{bmatrix}\right )^2
+ \left (M_{(i,j)}(I) - \begin{bmatrix}
                            i \\
                            j
                        \end{bmatrix}\right)^2} \right]
\end{equation}

\section {Experiments}
\subsection{Dataset}
We evaluate our model on the Caltech-UCSD Birds-200-2011 \cite{WahCUB_200_2011} dataset which has 200 classes of birds with 5,994 train images and 5,794 test images. We follow the same splits as in \cite{FGR} for our experiments and report the top-1 classification accuracy. 

\subsection{Implementation Details}
We use $448\times448$ input image size in all experiments. All experiments are run for 110 epochs using stochastic gradient descent (SGD) optimizer with a momentum of 0.9, batch size 8 and learning rate 0.001, which decays by 0.1 after 50 epochs. Random crop and center crop augmentations are used during training and testing respectively. For deconstructing images using RCM, we use patch size of $7 \times 7$. Our implementation is based on PyTorch \cite{PyTorch} with one Quadro RTX 6000 GPU.

\subsection{Baseline Formation}
Our baseline consists of ResNet-50 backbone followed by a 200-class classification head and achieves $86.36\%$ top-1 test accuracy on CUB dataset. We perform extensive experiments to construct a strong baseline for the subsequent SSL based FGVC task.

\begin{table} [h!]
\parbox{.4\linewidth}{
\centering
\begin{tabular}{c | c}
\textbf{Batch Size} & \textbf{Test Accuracy}  \\
\hline
32         & 80.66          \\
16         & 83.67          \\
\textbf{8}          & \textbf{85.19}          \\
4          & 84.36          \\
2          & 75.35         
\end{tabular}
\caption{Effect of changing the batch size on ResNet-50 accuracy. Batch size of 8 leads to better accuracy on CUB dataset. Only Random Horizontal Flip transformation is used during training.}
\label{batch_size}
}
\hspace{0.2cm}
\parbox{.575\linewidth}{
\centering
\begin{tabular}{l | c | c}
\textbf{Model}               & \textbf{Transform} & \textbf{Test Accuracy}  \\
\hline
ResNet-50           & HF             & 85.19          \\
\textbf{ResNet-50}           & \textbf{+ C. Crop}   & \textbf{86.36}          \\
\hline
ResNet-50 + DB      & + C. Crop   & 86.00          \\
ResNet-50 + GB      & + C. Crop   & 86.17          \\
ResNet-50 + DB + GB & + C. Crop   & 86.27         
\end{tabular}
\caption{Effect of introducing additional data augmentation, diversification block (DB) and gradient-boosting (GCE) loss to the ResNet-50 model. HF stands for Horizontal Flip and C. Crop stands refers to Center Crop transformation. The batch size of 8 is used in all experiments. The standard ResNet-50 achieves the best test accuracy. \label{db_gb}}
}
\vspace{-5mm}
\end{table}

\paragraph{Batch Size:} Table. \ref{batch_size} shows the FGVC accuracy at different values of batch size. A ResNet-50 model is trained in all the experiments with horizontal flip transformation randomly applied on the training batches. Decreasing the batch size increases the accuracy and we reach to the best accuracy of $85.19\%$ at batch size $8$. This is due to the fact that, large-batch methods tend to converge to sharp minima and generalize less, especially in the case of small datasets like CUB-200-2011 \cite{DBLP:journals/corr/KeskarMNST16, DBLP:journals/corr/abs-1804-07612}. Decreasing the batch size further (i.e. 4, 2) makes the model capture more noise and hence decreases the test accuracy. The results can be different for large datasets like ImageNet \cite{russakovsky2015imagenet}, where we may use a relatively larger batch size to reduce the training noise and improve the generalization.

\begin{figure}[h!]
\centering
{\includegraphics[width=0.176\textwidth]{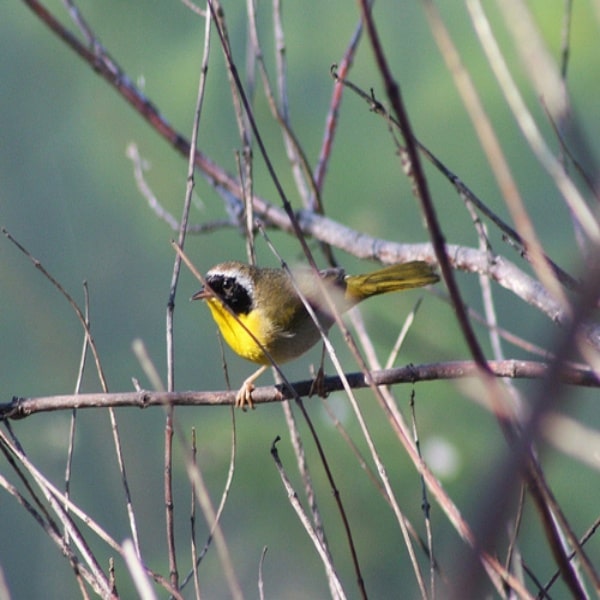}}
\hfill
{\includegraphics[width=0.176\textwidth]{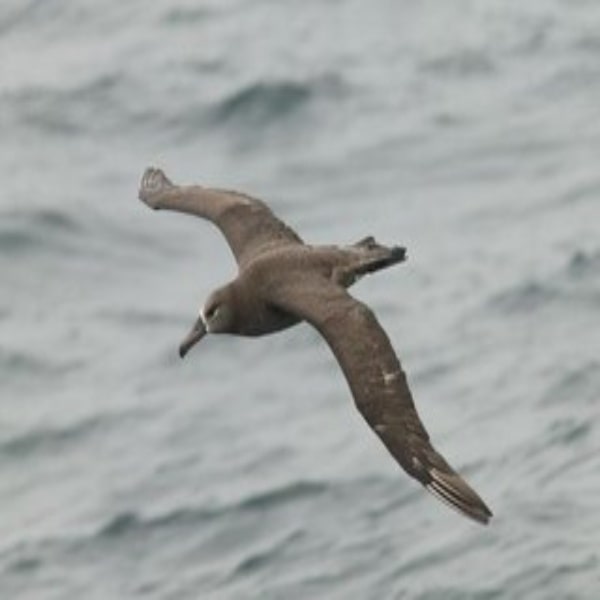}}
\hfill
{\includegraphics[width=0.176\textwidth]{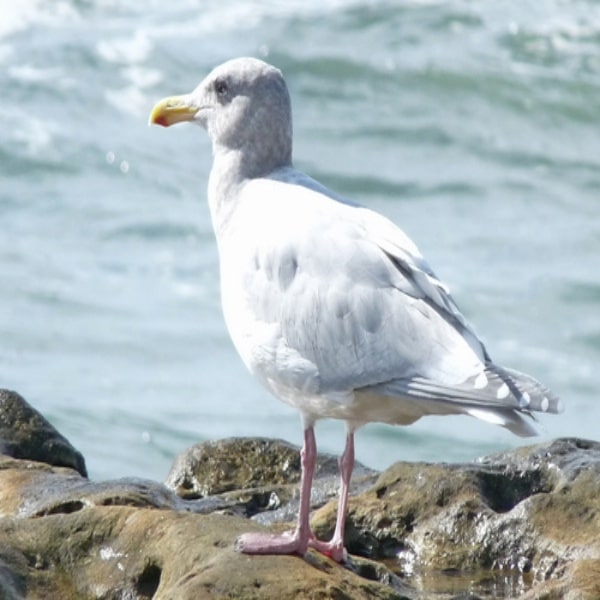}}
\hfill
{\includegraphics[width=0.176\textwidth]{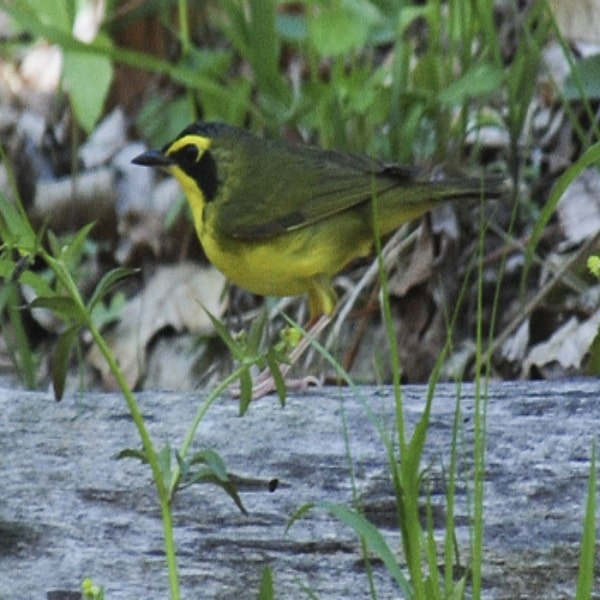}}
\hfill
{\includegraphics[width=0.176\textwidth]{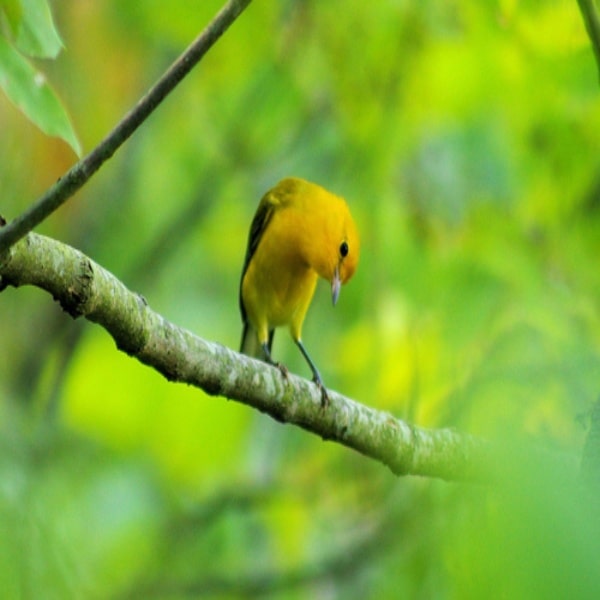}}
\caption{The figure shows a few samples (resized to $600 \times 600$) of the CUB dataset from different classes. Note that, mostly the birds are located at the center of the image.}
\label{cub_samples}
\end{figure}

\paragraph{Choice of Transformations:}
We found that the choice of data transformation during training and testing greatly effects the FGVC accuracy. As shown in Table. \ref{db_gb}, using random crop transformation during training and center crop transformation during testing improves the baseline accuracy from $85.19\%$ to $86.36\%$. The input image is resized to $600 \times 600$ pixels and then center cropped to $448 \times 448$ pixels during testing. This increase in accuracy could be because of the fact that most of the birds in the CUB dataset are located roughly at the center of the image.  Performing center cropping as a preprocessing allows the model to get a full view of the bird (Figure. \ref{cub_samples}).

\paragraph{Diversification Block and Gradient-boosting Loss:}
We use the concept from \cite{FGR} and introduce DB and GCE loss to the baseline and found that it does not contribute any improvement. 
Note that we try out different peak ($P_{peak}$) and patch ($P_{patch}$) suppression probabilities along with different suppressing factors ($\alpha$). We achieved the best results using 
$P_{peak}=0.2$, $P_{patch}=0.2$, $\alpha=0.5$ and patch size $(K)=2$. The DB is used only during the training phase. In order to capture the discriminative features, we train the network using CE loss for the first 10 epochs and then shift to GCE loss for another 100 epochs to focus on the difficult classes. 
Adding DB to ResNet-50 drops the test accuracy by $0.36\%$ while the reduction in accuracy with the use of GCE is $0.19\%$. Using both DB and GCE results in $86.27\%$ test accuracy which is only $0.09\%$ less than that of the baseline.

\subsection{Rotation as SSL Pretext Task}
In our experiments, we try out different $\lambda$ values to study the effect of self-supervision on the performance of the model. Table. \ref{diff_lamda} shows the effect of changing the rotation loss weight $(\lambda)$ value on the accuracy of the model and it is evident that increasing the $\lambda$ decreases the accuracy. This indicates that by increasing $\lambda$, the model focuses more on capturing the generic features for predicting rotation and deviates the model from capturing the subtle features, which are of prime importance for fine-grained recognition task.


\begin{center}
\begin{tabular}{c | c }

\textbf{$\lambda$} & \textbf{Test Accuracy} \\
\hline
\textbf{Baseline FGVC} ($\lambda$ = 0) & \textbf{86.36} \\
0.1       & 85.20   \\
0.3       & 84.58   \\
0.5       & 84.03   \\
0.7       & 82.55   \\
\end{tabular}
\captionof{table}{Effect of adding rotation as an auxiliary self-supervised task to the baseline for different values of $\lambda$ is shown in this table. Here $\lambda$ is the rotation loss weight that decides the contribution of rotation loss to the overall loss. We note that increasing $\lambda$ decreases the accuracy. \label{diff_lamda}}
\end{center}


\vspace{-2mm}
\subsection{Contrastive Self-supervised Learning}
Following the exceptional performance of contrastive SSL techniques on many downstream tasks, we study the usefulness of a few recent CL techniques for the fine-grained classification task.

\begin{table} [h!]
\parbox{.4\linewidth}{
\centering
\begin{tabular}{l | c}
\textbf{Method}              & \textbf{Test Accuracy}  \\
\hline
Supervised          & \textbf{86.36}          \\
\hline
PIRL                & 80.16          \\
SimCLR              & 75.31          \\
SwAV                & 77.44          \\
\textbf{Barlow Twins}        & \textbf{82.11}         
\end{tabular}
\caption{Effect of different weights initialization on the test accuracy of the baseline model. The pretrained weights are generated by training on ImageNet-1k dataset. For SSL methods, we use the pretrained weights from VISSL \cite{goyal2021vissl}.}

\label{cl_initialization}
}
\hspace{0.2cm}
\parbox{.575\linewidth}{
\centering
\begin{tabular}{c | c | c | c}
\textbf{Resize Dims.} & \textbf{Patch Size} & \textbf{Crops} & \textbf{Test Accuracy}  \\
\hline
$255 \times 255$ & $64 \times 64$   & 9            & 82.30          \\
\hline
$384 \times 384$ & $128 \times 128$ & 9            & 84.70          \\
$768 \times 768$ & $256 \times 256$ & 9            & 85.15          \\
$1024 \times 1024$ & $256 \times 256$ & \textbf{4}            & \textbf{85.65}         
\end{tabular}
\caption{Effect of introducing self-supervised PIRL \cite{misra2020self} loss to the baseline model for different jigsaw parameters. The resized images (of dimensions specified by Resize Dims.) are used for extracting jigsaw patches. The first row shows the standard jigsaw settings used in \cite{misra2020self}. For the last row, after resizing the image is center cropped to $512 \times 512$ before extracting patches.\label{cl_ssl}}
}
\vspace{-5mm}
\end{table}

\paragraph{Transfer Learning:}
Table. \ref{cl_initialization} shows the results of transfer learning to the CUB classification task using different CL-based self-supervised methods, where the pretext task has been performed on ImageNet dataset \cite{russakovsky2015imagenet}. The pretrained weights from VISSL \cite{goyal2021vissl} have been used to initialize all the experiments. Barlow Twins \cite{zbontar2021barlow} outperform its counterparts by achieving $82.11\%$ top-1 test accuracy which is only $4.25\%$ less as compared to the supervised pretraining. PIRL \cite{misra2020self} also achieves promising test accuracy of $80.16\%$ while SimCLR \cite{chen2020simple} and SwAV \cite{caron2020unsupervised} performs relatively poor as compared to Barlow Twins and PIRL. The experiments indicates the importance of weight initialization for the FGVC task where the accuracy gap is still significant compared to supervised pretraining unlike other standard downstream tasks (e.g. VOC Object Detection) as discussed in \cite{caron2020unsupervised, grill2020bootstrap}.

\paragraph{Contrastive SSL as an Auxiliary Task:}
Table. \ref{cl_ssl} shows the results of using the PIRL \cite{misra2020self} as an auxiliary task for FGVC classification. In standard settings, the original image is resized to $255 \times 255$ resolution and nine jigsaw patches of size $64 \times 64$ have been extracted from it. The representation generated from these patches along with the representation obtained from the original image are used to calculate the contrastive self-supervised loss. This SSL loss has been added to the supervised cross-entropy loss which gives us total loss for performing model training.
\begin{align*}
	& \mathcal{L}_{total} = \mathcal{L}_{classification} + \mathcal{L}_{PIRL}
\end{align*}

Following the standard setting (Table. \ref{cl_ssl} first row), our model achieves $82.30\%$ top-1 classification accuracy which is $4.06\%$ less as compared to baseline. Increasing the patch size to $128 \times 128$ and $256 \times 256$ increases the accuracy to $84.70\%$ and $85.15\%$ respectively. Introducing a resizing step followed by center crop transformation to the original image before extracting the patch images increases the model accuracy to $85.65\%$. Moreover, we used four patches instead of nine in this particular experiment, which also contributed to this improvement. Note that this accuracy score is still less by $0.71\%$ as compared to the baseline model which indicates the inefficiency of PIRL as an auxiliary task for CUB classification.

\subsection{Deconstruction and Construction Learning}
 In our experiments, we form the original DCL implementation \cite{chen2019destruction} as our baseline. In their work, the ground truths of jigsaw locations for the reconstruction are generated by measuring the similarity between the shuffled and the original patches. This method would be ambiguous in certain situations such as low texture variations in patches. We optimized ground truth generation by tracking the movement of each patch during the jigsaw shuffle. This improved the DCL performance from $86.90\%$ to an accuracy of $87.29\%$ (refer Table. \ref{dcl-results}). Another modification we adopted was to use MSE loss for reconstruction loss ($\mathcal{L}_{loc}$) instead of an L1 loss, which was originally proposed in \cite{chen2019destruction}. We also experimented using a classification head to predict the locations for reconstruction with a sigmoid activation and binary cross entropy loss (BCE), and multi label margin loss (MLML). As shown in Table. \ref{dcl-results}, the regression setting with MSE loss reportedly provide consistent best performance. 
 
 The effect of the application of DCL over the supervised model indicates the significance of self supervised learning in improving the performance of the model for the FGVC problem. We also perform an ablation study on the effect of the three loss function components to attain a deeper understanding of each. In these experiments, we keep the regression setting with MSE loss fixed. Table. \ref{dcl-ablation} indicates the results of the ablation study. 
 
\begin{table} [h!]
\parbox{.50\linewidth}{
\centering
\begin{tabular}{l | l | c }
\textbf{Model} & \textbf{Loss Fun.} & \textbf{Accuracy} \\
\hline
Original DCL & L1 & 86.90          \\
\hline
Improved DCL & L1 & 87.29    \\
\textbf{DCL with regression} & \textbf{MSE} & \textbf{87.41} \\
DCL with class & MLML & {87.17} \\
DCL with class & BCE & {87.21} \\   
\end{tabular}
\caption{Effect of various modifications on the DCL implementation. Improved DCL refers to training with optimized ground truth generated for jigsaw locations. Experiments with reconstructed jigsaw locations as regression and classification with various loss functions are shown. The regression setting with MSE loss outperforms the others.}
\label{dcl-results}
}
\hspace{0.3cm}
\parbox{.45\linewidth}{
\centering
\begin{tabular}{l | c }
\textbf{DCL with regression} & \textbf{Accuracy}   \\
\hline
Supervised(ResNet-50) & 86.36 \\ \hline
 $\boldsymbol{\mathcal{L}_{cls} + \mathcal{L}_{adv}+ \mathcal{L}_{loc}}$ & \textbf{87.41} \\
 $\mathcal{L}_{cls} + \mathcal{L}_{loc}$ & 87.26 \\
 $\mathcal{L}_{cls} + \mathcal{L}_{adv}$ & 87.21 \\
 $\mathcal{L}_{cls} $ & {87.02}      
\end{tabular}
\caption{Effect of introducing deconstruction and construction learning self supervision over supervised pipeline, indicates significance of learning from local regions. An ablation study of the classification loss, adversarial loss and location loss in DCL is shown on the regression setting with MSE loss.}
\label{dcl-ablation}
}
\vspace{-5mm}
\end{table}

\subsection{Semi-supervised Learning}
We study the effect of introducing supervision to the baseline model in the semi-supervised setting where the model is trained using only $10\%$, $30\%$ and $50\%$ of the training data. The whole test dataset has been used during the evaluation of the model (Table. \ref{semi-supervised}). Adding rotation or PIRL based self-supervision does not improve the baseline benchmarks, contemporary, DCL outperforms the baseline in both top-1 and top-2 accuracy metrics. For top-1 accuracy, the performance gap between DCL and baseline is $17.08\%$ when using $10\%$ labels while the gap is only $5.42\%$ in case of $30\%$ labels. This emphasizes the efficacy of self-supervision under the limited data availability. 

\vspace{-2mm}
\begin{table}[ht!]
\centering
\begin{tabular}{c | c c c | c c c}
         & \multicolumn{3}{c |}{\textbf{Top-1}} & \multicolumn{3}{c}{\textbf{Top-2}}  \\
\hline
         & \textbf{10\%}  & \textbf{30\%}  & \textbf{50\%}      & \textbf{10\%}  & \textbf{30\%}  & \textbf{50\%}       \\
\hline
Baseline & 29.05 & 70.58 & 79.79     & 40.39 & 81.79 & 88.27      \\
\hline
Rotation & 20.22 & 63.43 & 76.11     & 28.12 & 75.52 & 85.24      \\
PIRL     & 23.94 & 68.20 & 78.91     & 34.79 & 79.06 & 88.34      \\
\textbf{DCL}      & \textbf{46.13}      & \textbf{76.00}      &    \textbf{82.12}       &  \textbf{59.22}     & \textbf{85.13}      & \textbf{90.00}          
\end{tabular}
\caption{Comparison of the top-1 and top-2 test accuracies of different methods in semi-supervised setting. $10\%$, $30\%$ and $50\%$ of the training dataset has been used for the training while $100\%$ test dataset has been used during testing. DCL outperforms other methods in all categories.}
\label{semi-supervised}
\vspace{-5mm}
\end{table}

On an abosolute scale, DCL achieves $82.12\%$ top-1 accuracy by using only $50\%$ training data. This is only $4.24\%$ less as compared to the baseline model with $100\%$ training data. This is a great balance between accuracy and annotation cost where we achieve $82.12\%$ accuracy reducing the annotation cost to half. Furthermore, the performance gap between top-1 and top-2 accuracies is significant, indicating the confusion mainly happens between the top 2 classes for the fine-grained CUB classification task.

\section{Discussion}
This work studies the effectiveness of various self supervision techniques for fine-grained classification problem. Specifically, the usefulness of of rotation as pretext task, contrastive learning and DCL has been explored on the CUB FGVC problem. FGVC problems aims to distinguish subordinate level categories within a general category and therefore would have little variance between its classes. Moreover, the high intra-class variations in the data makes the problem more challenging. The choice of self supervision in this problem setting, depends on its efficiency in promoting learning in the model, to capture the fine grained details in order to distinguish between the very similar classes. Rotation as an auxiliary task promotes the model to learn global feature representations from the images, and diverts it from focusing on subtle features. The experiments prove its inefficiency for this problem setting. However, contrastive learning methods like PIRL that uses jigsaw patches to learn transformation-invariant representations from the images attempts to focus on discriminative local regions, but struggles to accurately localize them. 
\begin{figure}[h!]
\centering
\includegraphics[width=\linewidth]{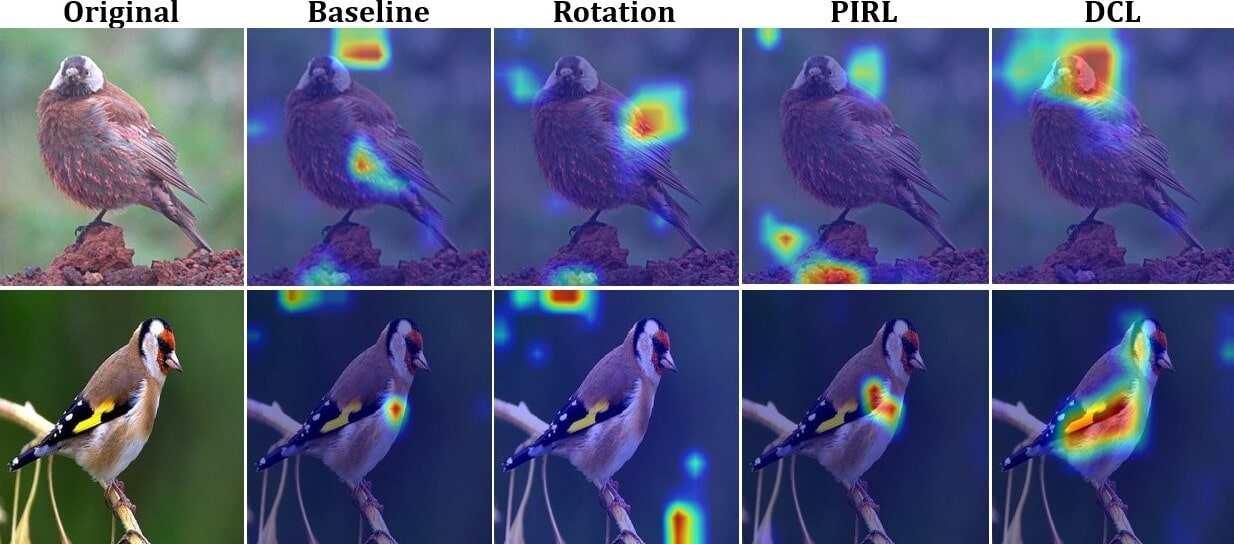}
\caption{The figure shows the Grad-CAM visualization of the various SSL methods experimented on the FGVC problem. The discriminating attributes such as the crown of the gray-crowned rosy finch (shown in top row) has been successfully identified and localized by DCL. Similarly, the red face and yellow patched wings are some discriminative features of the European gold finch (shown in bottom row), which has also been correctly localized.}
\label{combined_cam}
\end{figure}

The self supervision with DCL on the other hand, succeeded in capturing local discriminative features. The deconstruction learning forces the model to focus on local object parts and hence learns to capture discriminative subtle features, while reconstruction helps in learning the correlation between the parts. Gradient-weighted class activation map (Grad-CAM) \cite{selvaraju2017grad} visualizations of the various self supervised methods shown in Figure. \ref{combined_cam} supports the experimental findings. We believe that SSL tasks such as colorization and exploratory analysis for localization on PIRL, as strong future directions for our work.

\section{Conclusion}
The paper studies the effectiveness of SSL for fine-grained classification task on CUB-200-2011 dataset. Our baseline comprising a ResNet-50 backbone achieves $86.36\%$ top-1 classification accuracy. The various SSL auxiliary tasks we explored in this work are rotation, contrastive learning with PIRL, and DCL. The experiments conducted for this work illustrates that, SSL techniques that enhances the ability of the model to identify discriminative local object parts, are the best fit for the FGVC problem. Therefore, self-supervision with rotation that promotes learning semantic representations does not prove to be helpful in improving classification of such a fine-grained dataset. However, self-supervision such as PIRL that learns representations from local jigsaw regions showcase better learning comparatively. Furthermore, learning from deconstructed jigsaw patches and reconstruction using DCL allows the model to capture discriminative subtle features, surpassing the performance of the baseline model with an accuracy of $87.41\%$. Upon the Grad-CAM visualizations of the various methodologies, we understand that colours of the subtle parts of the birds are strong discriminative features that contribute to accurate classification. Following the observation, we propose introducing self-supervision using colorization as a potential future direction for this work.

\clearpage
\bibliography{references}
\clearpage
\begin{appendices}

\section{Diversification Block} \label{DB}
To attain high accuracy in a FGVC problem, the model must learn to capture the subtle features from the data in order to distinguish between the very similar classes. The model must therefore be forced to learn these subtle features, rather than being let to easily capture the most obvious discriminative features. To achieve this, we have adopted the diversification block introduced in \cite{FGR} which hides some of the discriminative features using a two stage suppression technique. The suppressions are applied to the CAMs denoted by $A$, from the model corresponding to the $C$ classes of the dataset. The suppression is applied in a pixel level and at patch level by combining the binary masks generated at the two levels. The overall suppression mask denoted by M, has the same dimensions of the CAMs, such that $M \in \R^{C \times H \times W}$ where $C$ represents the number of classes, $H$ and $W$ representing the height and width of the activation maps of the last convolution layer.

\begin{figure}[h!]
\centering
\includegraphics[width=\linewidth]{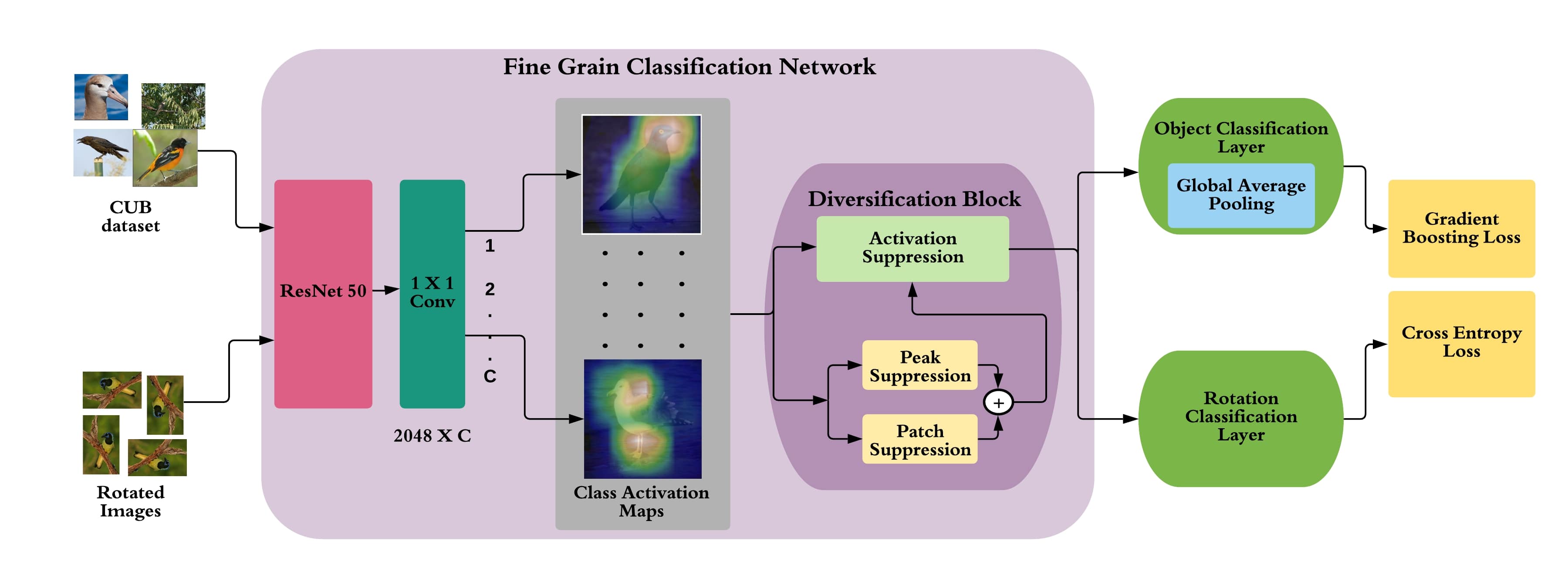}
\caption{Overview of our overall architecture. The model uses joint supervision from a supervised fine grained recognition and pretext rotation classification task. A resnet-50 backbone extracts feature and the class specific activations from the $1 \times 1$ convolution layer is passed to the diversification block. The suppressed activations are then passed to object classifier with gradient boosting loss and rotation classifier with cross entropy loss.}
\label{fig:blockdgm}
\end{figure}

\textbf{Peak suppression:} At the pixel level, a mask $M_{peak}$ is generated to suppress the peak values of the CAMs. Intuitively, this indicates that with peak suppression, we hide the most discriminative features captured by the model as the peaks of the CAMs are the regions that contribute majorly towards the classification. This is done by finding the peak locations of the CAMs and setting the values at the corresponding locations of $M_{peak}$ as 1, where a value 1 indicates locations to be suppressed. The $M_{peak}$ mask is then multiplied with a Bernoulli random variable $r_c$ of probability $P_{peak}$. This introduces a randomness to the peak suppression, where the values of $P_{peak}$ is a hyperparameter choice. The mask per class $M_{c \ peak}$ is formed by first forming a mask $P_c$ and then element wise multiplying with the random variable $r_c$.

\begin{equation*}
     P_c(i,j) =
    \begin{cases}
      1, & \text{if}\ A_c(i,j) = Max(A_c) \\
      
      0, & \text{otherwise}
    \end{cases}
\end{equation*}

$$
M_{c \ peak} = r_c * P_c \ , \ where \ r_c \in [0,1] \sim \textrm{Bernoulli}(P_{peak})
$$

\textbf{Patch suppression:} As we have suppressed the most discriminative features in the pixel level, some of the other mildly discriminative features are suppressed at the patch level. The CAMs are first split into patches of fixed size $K$, giving $l \times m$ patches of $K\times K$ sizes, denoted by $M_{patch}[l,m]$, where $l = H/K$ and $m = W/K$. These patches are then randomly suppressed with a probability $P_{patch}$ using a mask $M_{patch}$ by setting the corresponding patch locations of the mask as 1. Here the values of $P_{patch}$ is also a hyperparameter choice.
$$M_{c \ patch} = \{M_{c\ patch}[l,m] \in [0,1] \sim \textrm{Bernoulli}(P_{patch})\}$$

To ensure that we do not suppress the peak locations at this stage, the values at the corresponding locations of peaks in mask $M_{patch}$ is set to 1.
$$ M_{c \ patch}(i,j) = 0 \  \text{if}\ A_c(i,j) = Max(A_c)  $$

The two patches are then combined to form the cam suppression mask $M_c$ corresponding to each class. This is then used for suppressing the discriminative features of the CAMs by scaling the values of the CAMs at the corresponding locations of 1 in the mask $M_c$ by a scaling or suppression factor $\alpha$. Here $\alpha$ is a hyperparameter choice that controls the degree of suppression. This scaling suppresses the activations at the discriminative regions thus causing the subtle features to contribute towards the logits and therefore forcing the model to learn these features.
$$ M_c = M_{c \ peak} + M_{c \ patch}$$
\begin{equation*}
     A_c'(i,j) =
    \begin{cases}
      \alpha * A_c(i,j), & \text{if}\ M_c(i,j) = 1 \\
       A_c(i,j), & \text{if}\ M_c(i,j) = 0 
    \end{cases}
\end{equation*}

\section{Gradient-boosting Cross Entropy Loss}\label{GCE}
Gradient-boosting cross entropy (GCE) loss  proposed in \cite{FGR} makes the model focus on improving the classification of confusing classes by magnifying the gradient updates. Irrespective to standard cross entropy (CE) loss, which simply considers all negative classes equally, the GCE only focuses on top k negative classes which results in boosting the gradients. The eq. \ref{gce} shows the GCE loss calculation. Here $l$ is the ground truth label, $s$ are the confidence scores and $J^\prime$ is the set containing top $k$ classes.
\begin{equation} \label{gce}
    GCE(s, l) = - \log \left ( \frac{\exp(s_l)}{\exp(s_l) + \sum_{i \in J^\prime} \exp(s_i)} \right )
\end{equation}

\end{appendices}

\small

\end{document}